\documentclass[sigconf]{acmart}

\usepackage{balance} 

\usepackage{breakurl}

\setcopyright{none}

\acmYear{2018} 
\copyrightyear{2018} 
\acmConference[FATREC'18]{Workshop on Responsible Recommendation}{October 6, 2018}{Vancouver, Canada}


\begin{document}

\title[Explainable recommendations for fair lending]{Fair lending needs explainable models for responsible recommendation}

\author{Jiahao Chen}
\affiliation{%
    \institution{Capital One}
    \streetaddress{11 W 19th St}
    \city{New York}
    \state{New York}
    \postcode{10011-4218}
    \country{USA}
}
\email{jiahao.chen@capitalone.com}
\orcid{0000-0002-4357-6574}

\begin{abstract}
The financial services industry has unique explainability and fairness challenges arising from compliance and ethical considerations in credit decisioning. These challenges complicate the use of model machine learning and artificial intelligence methods in business decision processes.
\end{abstract}

\maketitle

Financial services companies in the USA make highly regulated business decisions that complicate the use of recommender systems and other forms of artificial intelligence. Key business processes such as determining who qualifies for lines of credit (personal loans, credit cards, mortgages, etc.) must be shown to comply with fair lending laws such as the
Equal Credit Opportunity Act (ECOA) \cite{ecoa},
Fair Credit Reporting Act (FCRA) \cite{fcra},
Fair and Accurate Credit Transactions Act (FACTA) \cite{facta},
Fair Housing Act (FHA) \cite{fha},
Fair and Equal Housing Act (FEHA) \cite{feha},
and
Consumer Credit Protection Act (CCPA) \cite{ccpa}.
Some of these laws define protected classes (see Table~\ref{table:classes}), for which discrimination on the basis of a customer's membership in those classes is prohibited.

Under these fair lending laws, lenders must demonstrate that their business decisions do \textbf{not} discriminate. However, there are multiple notions of discrimination in fair lending. The two major theories of discrimination are:

\begin{description}

\item[disparate treatment]\hspace{-1.3mm} \cite{Zimmer:1996aa}, informally, intentionally treating people differently on the basis on a protected class, and

\item[disparate impact]\hspace{-1.3mm} \cite{Rutherglen:1987aa}, informally, discriminating against any protected class as a resulting from implementing of a facially neutral policy.
\end{description}

Different theories of discrimination may apply for different laws.
Disparate impact as a theory of discrimination under FHA has been confirmed by the Supreme Court; however, there is some uncertainty as to whether disparate impact is a proper theory of discrimination under ECOA \cite{Cubita:2006aa,Avery:2012aa,Ropiequet:2014aa,McDonald2015:aa}.\footnote{While several courts and regulators believe that disparate impact should be considered under ECOA \cite{OCC:2017,CFPB:2018}, the Bureau of Consumer Financial Protection (CFPB) has signaled that it may issue guidance in the near future that will limit the use of disparate impact claims under ECOA \cite{CFPB:2018-05}.}

\begin{table}

\caption{\label{table:classes} Protected classes defined under US fair lending laws such as the Fair Housing Act (FHA)~\cite{fha} and Equal Credit Opportunity Act (ECOA)~\cite{ecoa}.}

\begin{tabular}{|l|c|c|}
\hline
Law & FHA\cite{fha} & ECOA\cite{ecoa} \\
\hline\hline
age & & X \\
color & X & X \\
disability & X & \\
exercised rights under CCPA\cite{ccpa} & & X \\
familial status (household composition) & X & \\
gender identity & X & \\
marital status (single or married) & & X \\
national origin & X & X \\
race  & X & X \\
recipient of public assistance & & X \\
religion & X & X \\
sex & X & X \\
\hline

\end{tabular}

\end{table}

This position paper summarizes some of the main compliance, explainability and fairness considerations arising out of regulated business decisioning in the financial services industry that pose unique challenges for using recommender systems.

\paragraph{Disparate impact considerations constrain the use of features that correlate strongly with protected classes.}

The need to demonstrate lack of disparate impact means that a model for credit risk has to avoid features like zip code, which is highly correlated with race \cite{Kevin:2006aa}, a protected class.
Using zip code in a model therefore runs the risk of redlining, the denial of services in neighborhoods populated mainly by racial minorities \cite{CohenCole:2011aa,OCC:2017,CFPB:2018}.
Other variables that may be predictive of credit risk, such as length of credit history, correlate with age of customer, another prohibited class \cite{Avery:2012aa}, and may require remediation in automated scoring systems \cite{CFPB:2018}.
Even seemingly innocuous policies like a minimum principal amount for a loan~\cite{Ritter:2012aa} may introduce bias against one or more protected classes~\cite{Aleo:2008aa,Steel:2010aa,ONeill:2016aa}.
Disparate impact considerations also complicate the use of nontraditional data sources such as social network data, which reflect and exacerbate deeply rooted societal inequalities \cite{Wei:2016aa,Hurley:2016aa,Barocas:2016aa}.

Similar considerations apply for models that are used in direct marketing to pre-screened potential customers \cite{Strader:2015aa,Butler:2016aa}, where credit bureau data are used to generate lists of consumers that qualify for a credit card offer.
According to FCRA, each marketing offer to a prescreened customer is a firm offer of credit; all a customer needs to do is accept the offer to obtain credit \cite{fcra}.
Therefore, marketing campaigns for credit have compliance considerations similar to credit decisioning models, on top of any issues around the consumers' perceptions of fairness \cite{Mayser:2011aa,Mayser:2013aa,Nguyen:2013aa}. Any artificial intelligence or machine learning models used for these purposes, including recommender systems, must therefore be capable of assessment for fair lending considerations.

\paragraph{Assessing discrimination poses unique challenges when customers' memberships in protected classes are unknown.}

Credit card companies do not generally collect information about an applicant's race, but may have a compliance need to demonstrate the lack of disparate impact with regard to race.
Regulators like the Consumer Financial Protection Bureau (CFPB) have published assessment methodologies that describe the use of proxy models to impute race labels \cite{Elliott:2009aa,CFPB:proxy2014}.
However, the resulting assessment seems to overestimate the amount of disparate impact \cite{Baines:2014aa,Zhang:2016aa} and has resulted in some controversy \cite{Baines:2014aa,Koren:2016aa,CFS:2015,CFS:2016,CFS:2017,CFPB:2018-05}.
To the best of our knowledge, assessing disparate impact in the absence of known labels is an area that has yet to be discussed in the fair, accountable and transparent machine learning community.

\paragraph{Fair lending laws require credit decisions to be explainable.}

An adverse action notice is required by ECOA and FCRA if a customer is denied credit based on information in a credit report~\cite{ecoa,fcra}.
According to the official interpretation of the law, such a notice must provide specific reasons for denying credit~\cite[\P 9(b)(2)]{regulationb}.
Therefore, AI/ML systems used to extend credit must also be able to provide the explanations necessary for adverse action.

Having explainable processes is important not only for determining regulatory compliance, but also to ``debug'' and redress errors in business practice, such as data quality errors in credit bureaux and other data sources \cite{Taylor:2013}.
Having ``debuggable'' credit lending practices is even more desirable when considering that 5\% of Americans have errors in their credit reports that adversely affect their creditworthiness \cite{FTC:2013aa}.
Accounting for such data quality errors has added moral and ethical dimensions \cite{Clarke:2016aa,Bayer:2018aa}, particularly when noting that the presence of such issues is itself correlated with protected classes, such as race, due to complex socioeconomic factors \cite{Ards:2001aa,Freeman:2013aa,Freeman:2017aa,Rothstein:2018aa}.
For example, Latinos/Hispanics are more likely than whites and African-Americans to have no credit history \cite{Ibarra:2007aa}.
Remediating such errors and omissions can break vicious cycles for people with bad credit due to erroneous credit reports, and for people who have no credit and hence no credit history \cite{ONeill:2016aa}.
Debuggability can also help businesses quickly identify problematic features that have the potential to generate customer dissatisfaction \cite{Lieber:2009aa}.

\paragraph{Can complex models be explained and proven to be fair on a case-by-case basis?}

The desire for explainability appears to clash with the desire to use complex models that may provide improved lift or classification accuracy \cite{Fernandez-Delgado:2014aa,Turner:2016aa}.
This kind of complexity has sometimes been termed ``opacity at scale'' \cite{Burrell:2016aa}.
Multi-armed bandits \cite{Vermorel:2005aa,Bubeck:2012aa} and other forms of reinforcement learning \cite{Kaebling:1996aa,Mnih:2015aa} pose further explainability challenges, due to the statefulness resulting from multiple rounds of learning.
However, recent work on automatic model explanations \cite{Turner:2016aa,lime} suggest that it is possible to generate relatively short explanations for the predictions produced for any particular set of inputs and outputs.
Inspired by these investigations, we propose to study how to generate explanations that are rated favorably by human subjects.
Possible avenues of study are stated below:

\paragraph{Hypothesis: different audiences require different explanations.}

Explainability cannot exist as a quality purely independent of a target audience.
A complete description of a business process that employs machine learning models requires mastery of the language of the business context in
addition to understanding the jargon and nuance surrounding statistical
modeling.
Only an expert data scientist who is intimately familiar with the
business has a reasonable chance of understanding the full operational
definition of an abstract concept like creditworthiness. Consequently,
explanations that are satisfactory to other audiences such as business
executives, data scientists in other industries or academia, lawmakers and
regulators, and customers, require different levels of distillation to convey
appropriate levels of explanation that use vocabularies and phrasing
comprehensible to the target audiences. We hypothesize that satisfaction with
explanations will be at best moderately correlated when the same explanation is
presented to domain experts and to the lay public.

\paragraph{Hypothesis: directly measurable features are more explainable.}

We argue that features that are directly measurable, such as a cutoff
based on a customer's annual salary, are more intuitive and hence
convey more explanatory power than a feature that mixes many such
features. Examples of the latter include being in the highest risk
twentile predicted by another model, or the largest principal component
built from all credit bureau information, or even a landmark feature
produced by running a fast (if less accurate) machine learning algorithm
\cite{Pfahringer:2000aa}. We posit that landmark features and principal
components, while popular as meta-features for meta-learning
\cite{Bardenet:2013aa,Feurer:2015aa}, are less intuitive and hence more
difficult to use in satisfactory explanations. Thus locality in the space of
directly measured features is an important factor that explains explainability.

\paragraph{Hypothesis: explanations from complex models are less satisfactory
than explanations from local approximants.}

Building upon the idea that locality in feature space promotes explainability,
we also speculate that explanations built from the output of local
approximants, such as Local Interpretable Model-Agnostic Explanations (LIME)
\cite{lime} or Quantitative Input Influence (QII) \cite{Datta:2016aa}, will be
rated as more explainable than explanations built directly upon the result of a
complex model such as a neural network.

\paragraph{Hypothesis: human ratings of satisfaction with explanations may differ from metrics based on statistical quality.}

We argue that the relationship between customer satisfaction with explanations
and metrics proposed in the literature, such as Turner's model quality score
\cite{Turner:2016aa}, is well worth investigating. We already know that in
other machine learning disciplines such as topic modeling that human
evaluations can behave in unexpected ways: in topic modeling, human ratings of
topic quality are negatively correlated with statistical likelihood measures
\cite{Chang:2009aa}. Furthermore, given our assertion above that explainability
cannot be independent of the target audience, we expect that general customers
will demand different level of explanations from domain experts.

\begin{acks}
I would like to thank Andrew Buemi, Carla Greenberg, Brian Larkin, and Brian Mills for their careful reading of the manuscript and their helpful suggestions and comments.
\end{acks}

\section*{Disclaimer}
The author of this paper is not a lawyer.
This paper does not constitute legal advice.
The positions herein are presented for the purpose of academic research discussions, and do not necessarily reflect the views of Capital One.

\bibliographystyle{ACM-Reference-Format}
\bibliography{bib}


\begin{thebibliography}{61}


\ifx \showCODEN    \undefined \def \showCODEN     #1{\unskip}     \fi
\ifx \showDOI      \undefined \def \showDOI       #1{#1}\fi
\ifx \showISBNx    \undefined \def \showISBNx     #1{\unskip}     \fi
\ifx \showISBNxiii \undefined \def \showISBNxiii  #1{\unskip}     \fi
\ifx \showISSN     \undefined \def \showISSN      #1{\unskip}     \fi
\ifx \showLCCN     \undefined \def \showLCCN      #1{\unskip}     \fi
\ifx \shownote     \undefined \def \shownote      #1{#1}          \fi
\ifx \showarticletitle \undefined \def \showarticletitle #1{#1}   \fi
\ifx \showURL      \undefined \def \showURL       {\relax}        \fi
\providecommand\bibfield[2]{#2}
\providecommand\bibinfo[2]{#2}
\providecommand\natexlab[1]{#1}
\providecommand\showeprint[2][]{arXiv:#2}

\bibitem[\protect\citeauthoryear{Aleo and Svirsky}{Aleo and Svirsky}{2008}]%
        {Aleo:2008aa}
\bibfield{author}{\bibinfo{person}{Michael Aleo} {and} \bibinfo{person}{Pablo
  Svirsky}.} \bibinfo{year}{2008}\natexlab{}.
\newblock \showarticletitle{Foreclosure Fallout: the banking industry's attack
  on disparate impact race discrimination claims under the {F}air {H}ousing
  {A}ct and the {E}qual {C}redit {O}pportunity {A}ct}.
\newblock \bibinfo{journal}{\emph{Public Law Interest Journal}}
  \bibinfo{volume}{18}, \bibinfo{number}{1} (\bibinfo{year}{2008}),
  \bibinfo{pages}{1--66}.
\newblock
\urldef\tempurl%
\url{https://www.bu.edu/pilj/files/2015/09/18-1AleoandSvirskyArticle.pdf}
\showURL{%
\tempurl}


\bibitem[\protect\citeauthoryear{Ards and Myers}{Ards and Myers}{2001}]%
        {Ards:2001aa}
\bibfield{author}{\bibinfo{person}{Sheila~D Ards} {and}
  \bibinfo{person}{Samuel~L Myers}.} \bibinfo{year}{2001}\natexlab{}.
\newblock \showarticletitle{The Color of Money: Bad Credit, Wealth, and Race}.
\newblock \bibinfo{journal}{\emph{American Behavioral Scientist}}
  \bibinfo{volume}{45} (\bibinfo{year}{2001}), \bibinfo{pages}{223--239}.
\newblock
\urldef\tempurl%
\url{https://doi.org/10.1177/00027640121957141}
\showDOI{\tempurl}


\bibitem[\protect\citeauthoryear{Avery, Brevoort, and Canner}{Avery
  et~al\mbox{.}}{2012}]%
        {Avery:2012aa}
\bibfield{author}{\bibinfo{person}{Robert~B. Avery},
  \bibinfo{person}{Kenneth~P. Brevoort}, {and} \bibinfo{person}{Glenn Canner}.}
  \bibinfo{year}{2012}\natexlab{}.
\newblock \showarticletitle{Does Credit Scoring Produce a Disparate Impact?}
\newblock \bibinfo{journal}{\emph{Real Estate Economics}} \bibinfo{volume}{40},
  \bibinfo{number}{s1} (\bibinfo{year}{2012}), \bibinfo{pages}{S65--S114}.
\newblock
\urldef\tempurl%
\url{https://doi.org/10.1111/j.1540-6229.2012.00348.x}
\showDOI{\tempurl}


\bibitem[\protect\citeauthoryear{Baines and Courchane}{Baines and Courchane}{4
  11}]%
        {Baines:2014aa}
\bibfield{author}{\bibinfo{person}{Arthur~P Baines} {and}
  \bibinfo{person}{Marsha~J Courchane}.} \bibinfo{year}{2014-11}\natexlab{}.
\newblock \bibinfo{title}{Fair Lending: Implications for the Indirect Auto
  Finance Market}.
\newblock
\newblock
\urldef\tempurl%
\url{https://www.crai.com/sites/default/files/publications/Fair-Lending-Implications-for-the-Indirect-Auto-Finance-Market.pdf}
\showURL{%
\tempurl}


\bibitem[\protect\citeauthoryear{Bardenet, Brendel, K{é}gl, and
  Sebag}{Bardenet et~al\mbox{.}}{2013}]%
        {Bardenet:2013aa}
\bibfield{author}{\bibinfo{person}{R{é}mi Bardenet}, \bibinfo{person}{Mátyás
  Brendel}, \bibinfo{person}{Balázs K{é}gl}, {and} \bibinfo{person}{Michèle
  Sebag}.} \bibinfo{year}{2013}\natexlab{}.
\newblock \showarticletitle{Collaborative Hyperparameter Tuning}. In
  \bibinfo{booktitle}{\emph{Proceedings of the 30th International Conference on
  International Conference on Machine Learning - Volume 28}}
  \emph{(\bibinfo{series}{ICML'13})}. \bibinfo{publisher}{JMLR.org},
  \bibinfo{pages}{II--199--207}.
\newblock
\urldef\tempurl%
\url{http://dl.acm.org/citation.cfm?id=3042817.3042916}
\showURL{%
\tempurl}


\bibitem[\protect\citeauthoryear{Barocas and Selbst}{Barocas and
  Selbst}{2016}]%
        {Barocas:2016aa}
\bibfield{author}{\bibinfo{person}{Solon Barocas} {and} \bibinfo{person}{Andrew
  Selbst}.} \bibinfo{year}{2016}\natexlab{}.
\newblock \showarticletitle{Big Data's Disparate Impact}.
\newblock \bibinfo{journal}{\emph{California Law Review}}
  \bibinfo{volume}{104}, \bibinfo{number}{1} (\bibinfo{year}{2016}),
  \bibinfo{pages}{671--729}.
\newblock
\urldef\tempurl%
\url{https://doi.org/10.15779/Z38BG31}
\showDOI{\tempurl}


\bibitem[\protect\citeauthoryear{Bayer, Ferreira, and Ross}{Bayer
  et~al\mbox{.}}{8 01}]%
        {Bayer:2018aa}
\bibfield{author}{\bibinfo{person}{Patrick Bayer}, \bibinfo{person}{Fernando
  Ferreira}, {and} \bibinfo{person}{Stephen~L. Ross}.}
  \bibinfo{year}{2018-01}\natexlab{}.
\newblock \showarticletitle{What Drives Racial and Ethnic Differences in
  High-Cost Mortgages? The Role of High-Risk Lenders}.
\newblock \bibinfo{journal}{\emph{The Review of Financial Studies}}
  \bibinfo{volume}{31}, \bibinfo{number}{1} (\bibinfo{year}{2018-01}),
  \bibinfo{pages}{175--205}.
\newblock
\urldef\tempurl%
\url{https://doi.org/10.1093/rfs/hhx035}
\showDOI{\tempurl}


\bibitem[\protect\citeauthoryear{Bubeck and Cesa-Bianchi}{Bubeck and
  Cesa-Bianchi}{2012}]%
        {Bubeck:2012aa}
\bibfield{author}{\bibinfo{person}{S{é}bastien Bubeck} {and}
  \bibinfo{person}{Nicol{ò} Cesa-Bianchi}.} \bibinfo{year}{2012}\natexlab{}.
\newblock \showarticletitle{Regret Analysis of Stochastic and Nonstochastic
  Multi-armed Bandit Problems}.
\newblock \bibinfo{journal}{\emph{Foundations and Trends in Machine Learning}}
  \bibinfo{volume}{5}, \bibinfo{number}{1} (\bibinfo{year}{2012}),
  \bibinfo{pages}{1--122}.
\newblock
\urldef\tempurl%
\url{https://doi.org/10.1561/2200000024}
\showDOI{\tempurl}


\bibitem[\protect\citeauthoryear{Burrell}{Burrell}{2016}]%
        {Burrell:2016aa}
\bibfield{author}{\bibinfo{person}{Jenna Burrell}.}
  \bibinfo{year}{2016}\natexlab{}.
\newblock \showarticletitle{How the machine `thinks': Understanding opacity in
  machine learning algorithms}.
\newblock \bibinfo{journal}{\emph{Big Data \& Society}} \bibinfo{volume}{3},
  \bibinfo{number}{1} (\bibinfo{year}{2016}), \bibinfo{pages}{1--12}.
\newblock
\urldef\tempurl%
\url{https://doi.org/10.1177/2053951715622512}
\showDOI{\tempurl}


\bibitem[\protect\citeauthoryear{Butler}{Butler}{6 08}]%
        {Butler:2016aa}
\bibfield{author}{\bibinfo{person}{Tammy Butler}.}
  \bibinfo{year}{2016-08}\natexlab{}.
\newblock \bibinfo{title}{Is your "target marketing" breaking fair lending
  laws?}
\newblock
\newblock
\urldef\tempurl%
\url{https://fairlendingdiversity.com/target-marketing-breaking-fair-lending-laws/}
\showURL{%
\tempurl}


\bibitem[\protect\citeauthoryear{Chang, Boyd-Graber, Gerrish, Wang, and
  Blei}{Chang et~al\mbox{.}}{2009}]%
        {Chang:2009aa}
\bibfield{author}{\bibinfo{person}{Jonathan Chang}, \bibinfo{person}{Jordan
  Boyd-Graber}, \bibinfo{person}{Sean Gerrish}, \bibinfo{person}{Chong Wang},
  {and} \bibinfo{person}{David~M. Blei}.} \bibinfo{year}{2009}\natexlab{}.
\newblock \showarticletitle{Reading Tea Leaves: How Humans Interpret Topic
  Models}. In \bibinfo{booktitle}{\emph{Proceedings of the 22nd International
  Conference on Neural Information Processing Systems}}
  \emph{(\bibinfo{series}{NIPS'09})}. \bibinfo{publisher}{Curran Associates
  Inc.}, \bibinfo{pages}{288--296}.
\newblock
\urldef\tempurl%
\url{https://papers.nips.cc/paper/3700-reading-tea-leaves-how-humans-interpret-topic-models}
\showURL{%
\tempurl}


\bibitem[\protect\citeauthoryear{Clarke}{Clarke}{2016}]%
        {Clarke:2016aa}
\bibfield{author}{\bibinfo{person}{Roger Clarke}.}
  \bibinfo{year}{2016}\natexlab{}.
\newblock \showarticletitle{Big data, big risks}.
\newblock \bibinfo{journal}{\emph{Information Systems Journal}}
  \bibinfo{volume}{26}, \bibinfo{number}{1} (\bibinfo{year}{2016}),
  \bibinfo{pages}{77--90}.
\newblock
\urldef\tempurl%
\url{https://doi.org/10.1111/isj.12088}
\showDOI{\tempurl}


\bibitem[\protect\citeauthoryear{Cohen-Cole}{Cohen-Cole}{1 05}]%
        {CohenCole:2011aa}
\bibfield{author}{\bibinfo{person}{Ethan Cohen-Cole}.}
  \bibinfo{year}{2011-05}\natexlab{}.
\newblock \showarticletitle{Credit card redlining}.
\newblock \bibinfo{journal}{\emph{Review of Economics and Statistics}}
  \bibinfo{volume}{93}, \bibinfo{number}{2} (\bibinfo{year}{2011-05}),
  \bibinfo{pages}{700--713}.
\newblock
\urldef\tempurl%
\url{https://doi.org/10.1162/REST_a_00052}
\showDOI{\tempurl}


\bibitem[\protect\citeauthoryear{{Committee on Financial Services}}{{Committee
  on Financial Services}}{5 11}]%
        {CFS:2015}
\bibfield{author}{\bibinfo{person}{{Committee on Financial Services}}.}
  \bibinfo{year}{2015-11}\natexlab{}.
\newblock \bibinfo{title}{Unsafe at Any Bureaucracy: {CFPB} Junk Science and
  Indirect Auto Lending}.
\newblock
\newblock
\urldef\tempurl%
\url{https://financialservices.house.gov/uploadedfiles/11-24-15_cfpb_indirect_auto_staff_report.pdf}
\showURL{%
\tempurl}


\bibitem[\protect\citeauthoryear{{Committee on Financial Services}}{{Committee
  on Financial Services}}{6 01}]%
        {CFS:2016}
\bibfield{author}{\bibinfo{person}{{Committee on Financial Services}}.}
  \bibinfo{year}{2016-01}\natexlab{}.
\newblock \bibinfo{title}{Unsafe at Any Bureaucracy, {Part II}: How the Bureau
  of Consumer Financial Protection Removed Anti-fraud Safeguards to Achieve
  Political Goals}.
\newblock
\newblock
\urldef\tempurl%
\url{https://financialservices.house.gov/uploadedfiles/cfpb_indirect_auto_part_ii.pdf}
\showURL{%
\tempurl}


\bibitem[\protect\citeauthoryear{{Committee on Financial Services}}{{Committee
  on Financial Services}}{7 01}]%
        {CFS:2017}
\bibfield{author}{\bibinfo{person}{{Committee on Financial Services}}.}
  \bibinfo{year}{2017-01}\natexlab{}.
\newblock \bibinfo{title}{Unsafe at Any Bureaucracy, Part {III}: The {CFPB}'s
  Vitiated Legal Case Against Auto-Lenders}.
\newblock
\newblock
\urldef\tempurl%
\url{https://financialservices.house.gov/uploadedfiles/1-18-17_cfpb_indirect_auto_staff_report_iii.pdf}
\showURL{%
\tempurl}


\bibitem[\protect\citeauthoryear{{Consumer Financial Protection
  Bureau}}{{Consumer Financial Protection Bureau}}{2014}]%
        {CFPB:proxy2014}
\bibfield{author}{\bibinfo{person}{{Consumer Financial Protection Bureau}}.}
  \bibinfo{year}{2014}\natexlab{}.
\newblock \bibinfo{title}{Using publicly available information to proxy for
  unidentified race and ethnicity: a methodology and assessment}.
\newblock
\newblock
\urldef\tempurl%
\url{https://www.consumerfinance.gov/data-research/research-reports/using-publicly-available-information-to-proxy-for-unidentified-race-and-ethnicity/}
\showURL{%
\tempurl}


\bibitem[\protect\citeauthoryear{{Consumer Financial Protection
  Bureau}}{{Consumer Financial Protection Bureau}}{8 03}]%
        {CFPB:2018}
\bibfield{author}{\bibinfo{person}{{Consumer Financial Protection Bureau}}.}
  \bibinfo{year}{2018-03}\natexlab{}.
\newblock \bibinfo{title}{{CFPB} Supervision and Examination Process}.
\newblock
\newblock
\urldef\tempurl%
\url{https://www.consumerfinance.gov/policy-compliance/guidance/supervision-examinations/}
\showURL{%
\tempurl}


\bibitem[\protect\citeauthoryear{{Consumer Financial Protection
  Bureau}}{{Consumer Financial Protection Bureau}}{8 05}]%
        {CFPB:2018-05}
\bibfield{author}{\bibinfo{person}{{Consumer Financial Protection Bureau}}.}
  \bibinfo{year}{2018-05}\natexlab{}.
\newblock \bibinfo{title}{Statement of the Bureau of Consumer Financial
  Protection on enactment of {S.J.} Res. 57}.
\newblock
\newblock
\urldef\tempurl%
\url{https://www.consumerfinance.gov/about-us/newsroom/statement-bureau-consumer-financial-protection-enactment-sj-res-57/}
\showURL{%
\tempurl}


\bibitem[\protect\citeauthoryear{Cubita and Hartmann}{Cubita and
  Hartmann}{2006}]%
        {Cubita:2006aa}
\bibfield{author}{\bibinfo{person}{Peter~N. Cubita} {and}
  \bibinfo{person}{Michelle Hartmann}.} \bibinfo{year}{2006}\natexlab{}.
\newblock \showarticletitle{The ECOA Discrimination Proscription and Disparate
  Impact---Interpreting the Meaning of the Words That Actually Are There}.
\newblock \bibinfo{journal}{\emph{The Business Lawyer}} \bibinfo{volume}{61},
  \bibinfo{number}{2} (\bibinfo{year}{2006}), \bibinfo{pages}{829--842}.
\newblock
\urldef\tempurl%
\url{http://www.jstor.org/stable/40688368}
\showURL{%
\tempurl}


\bibitem[\protect\citeauthoryear{Datta, Sen, and Zick}{Datta et~al\mbox{.}}{6
  05}]%
        {Datta:2016aa}
\bibfield{author}{\bibinfo{person}{A. Datta}, \bibinfo{person}{S. Sen}, {and}
  \bibinfo{person}{Y. Zick}.} \bibinfo{year}{2016-05}\natexlab{}.
\newblock \showarticletitle{Algorithmic Transparency via Quantitative Input
  Influence: Theory and Experiments with Learning Systems}. In
  \bibinfo{booktitle}{\emph{2016 IEEE Symposium on Security and Privacy (SP)}}.
  \bibinfo{pages}{598--617}.
\newblock
\urldef\tempurl%
\url{https://doi.org/10.1109/SP.2016.42}
\showDOI{\tempurl}


\bibitem[\protect\citeauthoryear{{Division of Consumer and Community
  Affairs}}{{Division of Consumer and Community Affairs}}{1 07}]%
        {regulationb}
\bibfield{author}{\bibinfo{person}{{Division of Consumer and Community
  Affairs}}.} \bibinfo{year}{2011-07}\natexlab{}.
\newblock \bibinfo{title}{12 {CFR} Supplement {I} to Part 202 - Official Staff
  Interpretations}.
\newblock
\newblock
\urldef\tempurl%
\url{https://www.law.cornell.edu/cfr/text/12/appendix-Supplement_I_to_part_202}
\showURL{%
\tempurl}


\bibitem[\protect\citeauthoryear{Elliott, Morrison, Fremont, McCaffrey,
  Pantoja, and Lurie}{Elliott et~al\mbox{.}}{9 04}]%
        {Elliott:2009aa}
\bibfield{author}{\bibinfo{person}{Marc~N. Elliott}, \bibinfo{person}{Peter~A.
  Morrison}, \bibinfo{person}{Allen Fremont}, \bibinfo{person}{Daniel~F.
  McCaffrey}, \bibinfo{person}{Philip Pantoja}, {and} \bibinfo{person}{Nicole
  Lurie}.} \bibinfo{year}{2009-04}\natexlab{}.
\newblock \showarticletitle{Using the Census Bureau's surname list to improve
  estimates of race/ethnicity and associated disparities}.
\newblock \bibinfo{journal}{\emph{Health Services and Outcomes Research
  Methodology}} \bibinfo{volume}{9}, \bibinfo{number}{2}
  (\bibinfo{year}{2009-04}), \bibinfo{pages}{69--83}.
\newblock
\urldef\tempurl%
\url{https://doi.org/10.1007/s10742-009-0047-1}
\showDOI{\tempurl}


\bibitem[\protect\citeauthoryear{{Federal Trade Commission}}{{Federal Trade
  Commission}}{3 02}]%
        {FTC:2013aa}
\bibfield{author}{\bibinfo{person}{{Federal Trade Commission}}.}
  \bibinfo{year}{2013-02}\natexlab{}.
\newblock \bibinfo{title}{In {FTC} Study, Five Percent of Consumers Had Errors
  on Their Credit Reports That Could Result in Less Favorable Terms for Loans}.
\newblock
\newblock
\urldef\tempurl%
\url{https://www.ftc.gov/news-events/press-releases/2013/02/ftc-study-five-percent-consumers-had-errors-their-credit-reports}
\showURL{%
\tempurl}


\bibitem[\protect\citeauthoryear{Fernández-Delgado, Cernadas, Barro, and
  Amorim}{Fernández-Delgado et~al\mbox{.}}{2014}]%
        {Fernandez-Delgado:2014aa}
\bibfield{author}{\bibinfo{person}{Manuel Fernández-Delgado},
  \bibinfo{person}{Eva Cernadas}, \bibinfo{person}{Senén Barro}, {and}
  \bibinfo{person}{Dinani Amorim}.} \bibinfo{year}{2014}\natexlab{}.
\newblock \showarticletitle{Do we Need Hundreds of Classifiers to Solve Real
  World Classification Problems?}
\newblock \bibinfo{journal}{\emph{Journal of Machine Learning Research}}
  \bibinfo{volume}{15} (\bibinfo{year}{2014}), \bibinfo{pages}{3133--3181}.
\newblock
\urldef\tempurl%
\url{http://jmlr.org/papers/v15/delgado14a.html}
\showURL{%
\tempurl}


\bibitem[\protect\citeauthoryear{Feurer, Springenberg, and Hutter}{Feurer
  et~al\mbox{.}}{2015}]%
        {Feurer:2015aa}
\bibfield{author}{\bibinfo{person}{Matthias Feurer},
  \bibinfo{person}{Jost~Tobias Springenberg}, {and} \bibinfo{person}{Frank
  Hutter}.} \bibinfo{year}{2015}\natexlab{}.
\newblock \showarticletitle{Initializing Bayesian Hyperparameter Optimization
  via Meta-learning}. In \bibinfo{booktitle}{\emph{Proceedings of the
  Twenty-Ninth {AAAI} Conference on Artificial Intelligence}}
  \emph{(\bibinfo{series}{AAAI'15})}. \bibinfo{publisher}{AAAI},
  \bibinfo{pages}{1128--1135}.
\newblock


\bibitem[\protect\citeauthoryear{Freeman}{Freeman}{2013}]%
        {Freeman:2013aa}
\bibfield{author}{\bibinfo{person}{Andrea Freeman}.}
  \bibinfo{year}{2013}\natexlab{}.
\newblock \showarticletitle{Payback: A structural analysis of the credit card
  problem}.
\newblock \bibinfo{journal}{\emph{Arizona Law Review}}  \bibinfo{volume}{55}
  (\bibinfo{year}{2013}), \bibinfo{pages}{151--199}.
\newblock


\bibitem[\protect\citeauthoryear{Freeman}{Freeman}{7 05}]%
        {Freeman:2017aa}
\bibfield{author}{\bibinfo{person}{Andrea Freeman}.}
  \bibinfo{year}{2017-05}\natexlab{}.
\newblock \showarticletitle{Racism in the Credit Card Industry}.
\newblock \bibinfo{journal}{\emph{North Carolina Law Review}}
  \bibinfo{volume}{95}, \bibinfo{number}{4} (\bibinfo{year}{2017-05}),
  \bibinfo{pages}{1071--1160}.
\newblock
\urldef\tempurl%
\url{http://scholarship.law.unc.edu/nclr/vol95/iss4/4}
\showURL{%
\tempurl}


\bibitem[\protect\citeauthoryear{Hurley and Adebayo}{Hurley and
  Adebayo}{2016}]%
        {Hurley:2016aa}
\bibfield{author}{\bibinfo{person}{Mikella Hurley} {and}
  \bibinfo{person}{Julius Adebayo}.} \bibinfo{year}{2016}\natexlab{}.
\newblock \showarticletitle{Credit Scoring in the Era of Big Data}.
\newblock \bibinfo{journal}{\emph{Yale Journal of Law \& Technology}}
  \bibinfo{volume}{18} (\bibinfo{year}{2016}), \bibinfo{pages}{148--216}.
\newblock
\urldef\tempurl%
\url{http://yjolt.org/credit-scoring-era-big-data}
\showURL{%
\tempurl}


\bibitem[\protect\citeauthoryear{Ibarra and Rodriguez}{Ibarra and
  Rodriguez}{2007}]%
        {Ibarra:2007aa}
\bibfield{author}{\bibinfo{person}{Beatriz Ibarra} {and} \bibinfo{person}{Eric
  Rodriguez}.} \bibinfo{year}{2007}\natexlab{}.
\newblock \showarticletitle{Latino Credit Card Use: Debt Trap or Ticket to
  Prosperity?}
\newblock \bibinfo{journal}{\emph{National Council of La Raza Issue Brief}}
  \bibinfo{volume}{17} (\bibinfo{year}{2007}).
\newblock


\bibitem[\protect\citeauthoryear{Kaebling, Littman, and Moore}{Kaebling
  et~al\mbox{.}}{6 05}]%
        {Kaebling:1996aa}
\bibfield{author}{\bibinfo{person}{Leslie~P Kaebling},
  \bibinfo{person}{Michael~L Littman}, {and} \bibinfo{person}{Andrew~W Moore}.}
  \bibinfo{year}{1996-05}\natexlab{}.
\newblock \showarticletitle{Reinforcement Learning: A Survey}.
\newblock \bibinfo{journal}{\emph{Journal of Artificial Intelligence Research}}
   \bibinfo{volume}{4} (\bibinfo{year}{1996-05}), \bibinfo{pages}{237--285}.
\newblock
\urldef\tempurl%
\url{https://jair.org/index.php/jair/article/view/10166}
\showURL{%
\tempurl}


\bibitem[\protect\citeauthoryear{Kevin and M.}{Kevin and M.}{2006}]%
        {Kevin:2006aa}
\bibfield{author}{\bibinfo{person}{Fiscella Kevin} {and}
  \bibinfo{person}{Fremont~Allen M.}} \bibinfo{year}{2006}\natexlab{}.
\newblock \showarticletitle{Use of Geocoding and Surname Analysis to Estimate
  Race and Ethnicity}.
\newblock \bibinfo{journal}{\emph{Health Services Research}}
  \bibinfo{volume}{41}, \bibinfo{number}{4p1} (\bibinfo{year}{2006}),
  \bibinfo{pages}{1482--1500}.
\newblock
\urldef\tempurl%
\url{https://doi.org/10.1111/j.1475-6773.2006.00551.x}
\showDOI{\tempurl}


\bibitem[\protect\citeauthoryear{Koren}{Koren}{6 08}]%
        {Koren:2016aa}
\bibfield{author}{\bibinfo{person}{James~Rufus Koren}.}
  \bibinfo{year}{2016-08}\natexlab{}.
\newblock \showarticletitle{Feds use {R}and formula to spot discrimination. The
  {GOP} calls it junk science}.
\newblock \bibinfo{journal}{\emph{Los Angeles Times}}
  (\bibinfo{year}{2016-08}).
\newblock
\urldef\tempurl%
\url{http://www.latimes.com/business/la-fi-rand-elliott-20160824-snap-story.html}
\showURL{%
\tempurl}


\bibitem[\protect\citeauthoryear{Lieber}{Lieber}{9 01}]%
        {Lieber:2009aa}
\bibfield{author}{\bibinfo{person}{Ron Lieber}.}
  \bibinfo{year}{2009-01}\natexlab{}.
\newblock \showarticletitle{American Express Kept a (Very) Watchful Eye on
  Charges}.
\newblock \bibinfo{journal}{\emph{The New York Times}}
  (\bibinfo{year}{2009-01}), \bibinfo{pages}{B1}.
\newblock
\urldef\tempurl%
\url{http://www.nytimes.com/2009/01/31/your-money/credit-and-debit-cards/31money.html?pagewanted=all}
\showURL{%
\tempurl}


\bibitem[\protect\citeauthoryear{Mayser}{Mayser}{1 07}]%
        {Mayser:2011aa}
\bibfield{author}{\bibinfo{person}{Sabine Mayser}.}
  \bibinfo{year}{2011-07}\natexlab{}.
\newblock \bibinfo{title}{Perceived Fairness of Differential Customer
  Treatment}.
\newblock
\newblock
\urldef\tempurl%
\url{https://doi.org/10.1177/1094670512464274}
\showDOI{\tempurl}


\bibitem[\protect\citeauthoryear{Mayser and von Wangenheim}{Mayser and von
  Wangenheim}{2013}]%
        {Mayser:2013aa}
\bibfield{author}{\bibinfo{person}{Sabine Mayser} {and}
  \bibinfo{person}{Florian von Wangenheim}.} \bibinfo{year}{2013}\natexlab{}.
\newblock \showarticletitle{Perceived Fairness of Differential Customer
  Treatment: Consumers' Understanding of Distributive Justice Really Matters}.
\newblock \bibinfo{journal}{\emph{Journal of Service Research}}
  \bibinfo{volume}{16}, \bibinfo{number}{1} (\bibinfo{year}{2013}),
  \bibinfo{pages}{99--113}.
\newblock
\urldef\tempurl%
\url{https://doi.org/10.1177/1094670512464274}
\showDOI{\tempurl}


\bibitem[\protect\citeauthoryear{McDonald}{McDonald}{2016}]%
        {McDonald2015:aa}
\bibfield{author}{\bibinfo{person}{Kevin~M McDonald}.}
  \bibinfo{year}{2015-2016}\natexlab{}.
\newblock \showarticletitle{Who's policing the financial cop on the beat? A
  call for judicial review of the {C}onsumer {F}Inancial {P}rotection
  {B}ureau's non-legislative rules}.
\newblock \bibinfo{journal}{\emph{Review of Banking \& Financial Law}}
  \bibinfo{volume}{35}, \bibinfo{number}{1} (\bibinfo{year}{2015-2016}),
  \bibinfo{pages}{224--271}.
\newblock
\urldef\tempurl%
\url{http://ssrn.com/abstract=2786093}
\showURL{%
\tempurl}


\bibitem[\protect\citeauthoryear{Mnih, Kavukcuoglu, Silver, Rusu, Veness,
  Bellemare, Graves, Riedmiller, Fidjeland, Ostrovski, Petersen, Beattie,
  Sadik, Antonoglou, King, Kumaran, Wierstra, Legg, and Hassabis}{Mnih
  et~al\mbox{.}}{5 02}]%
        {Mnih:2015aa}
\bibfield{author}{\bibinfo{person}{Volodymyr Mnih}, \bibinfo{person}{Koray
  Kavukcuoglu}, \bibinfo{person}{David Silver}, \bibinfo{person}{Andrei~A.
  Rusu}, \bibinfo{person}{Joel Veness}, \bibinfo{person}{Marc~G. Bellemare},
  \bibinfo{person}{Alex Graves}, \bibinfo{person}{Martin Riedmiller},
  \bibinfo{person}{Andreas~K. Fidjeland}, \bibinfo{person}{Georg Ostrovski},
  \bibinfo{person}{Stig Petersen}, \bibinfo{person}{Charles Beattie},
  \bibinfo{person}{Amir Sadik}, \bibinfo{person}{Ioannis Antonoglou},
  \bibinfo{person}{Helen King}, \bibinfo{person}{Dharshan Kumaran},
  \bibinfo{person}{Daan Wierstra}, \bibinfo{person}{Shane Legg}, {and}
  \bibinfo{person}{Demis Hassabis}.} \bibinfo{year}{2015-02}\natexlab{}.
\newblock \showarticletitle{Human-level control through deep reinforcement
  learning}.
\newblock \bibinfo{journal}{\emph{Nature}}  \bibinfo{volume}{518}
  (\bibinfo{year}{2015-02}), \bibinfo{pages}{529--533}.
\newblock
\urldef\tempurl%
\url{https://doi.org/10.1038/nature14236}
\showDOI{\tempurl}


\bibitem[\protect\citeauthoryear{Nguyen and Klaus}{Nguyen and Klaus}{2013}]%
        {Nguyen:2013aa}
\bibfield{author}{\bibinfo{person}{Bang Nguyen} {and}
  \bibinfo{person}{Philipp~Phil Klaus}.} \bibinfo{year}{2013}\natexlab{}.
\newblock \showarticletitle{Retail fairness: Exploring consumer perceptions of
  fairness towards retailers' marketing tactics}.
\newblock \bibinfo{journal}{\emph{Journal of Retailing and Consumer Services}}
  \bibinfo{volume}{20}, \bibinfo{number}{3} (\bibinfo{year}{2013}),
  \bibinfo{pages}{311--324}.
\newblock
\urldef\tempurl%
\url{https://doi.org/10.1016/j.jretconser.2013.02.001}
\showDOI{\tempurl}


\bibitem[\protect\citeauthoryear{{Office of the Comptroller of the
  Currency}}{{Office of the Comptroller of the Currency}}{7 01}]%
        {OCC:2017}
\bibfield{author}{\bibinfo{person}{{Office of the Comptroller of the
  Currency}}.} \bibinfo{year}{2017-01}\natexlab{}.
\newblock \bibinfo{title}{Credit Card Lending: {OCC} {C}omptroller's Handbook}.
\newblock
\newblock


\bibitem[\protect\citeauthoryear{O'Neill}{O'Neill}{2016}]%
        {ONeill:2016aa}
\bibfield{author}{\bibinfo{person}{Cathy O'Neill}.}
  \bibinfo{year}{2016}\natexlab{}.
\newblock \bibinfo{booktitle}{\emph{Weapons of Math Destruction: how big data
  increases inequality and threatens democracy}}.
\newblock \bibinfo{publisher}{Crown}.
\newblock
\urldef\tempurl%
\url{https://weaponsofmathdestructionbook.com}
\showURL{%
\tempurl}


\bibitem[\protect\citeauthoryear{Pfahringer, Bensusan, and
  Giraud-Carrier}{Pfahringer et~al\mbox{.}}{2000}]%
        {Pfahringer:2000aa}
\bibfield{author}{\bibinfo{person}{Bernhard Pfahringer}, \bibinfo{person}{Hilan
  Bensusan}, {and} \bibinfo{person}{Christophe~G. Giraud-Carrier}.}
  \bibinfo{year}{2000}\natexlab{}.
\newblock \showarticletitle{Meta-Learning by Landmarking Various Learning
  Algorithms}. In \bibinfo{booktitle}{\emph{Proceedings of the Seventeenth
  International Conference on Machine Learning}} \emph{(\bibinfo{series}{ICML
  '00})}. \bibinfo{publisher}{Morgan Kaufmann}, \bibinfo{pages}{743--750}.
\newblock
\urldef\tempurl%
\url{http://dl.acm.org/citation.cfm?id=645529.658105}
\showURL{%
\tempurl}


\bibitem[\protect\citeauthoryear{Ribeiro, Singh, and Guestrin}{Ribeiro
  et~al\mbox{.}}{6 08}]%
        {lime}
\bibfield{author}{\bibinfo{person}{Marco~Tulio Ribeiro},
  \bibinfo{person}{Sameer Singh}, {and} \bibinfo{person}{Carlos Guestrin}.}
  \bibinfo{year}{2016-08}\natexlab{}.
\newblock \showarticletitle{"Why Should {I} Trust You?": Explaining the
  Predictions of Any Classifier}. In \bibinfo{booktitle}{\emph{KDD '16
  Proceedings of the 22nd ACM SIGKDD International Conference on Knowledge
  Discovery and Data Mining}}. \bibinfo{publisher}{ACM},
  \bibinfo{pages}{1135--1144}.
\newblock
\urldef\tempurl%
\url{https://doi.org/10.1145/2939672.2939778}
\showDOI{\tempurl}


\bibitem[\protect\citeauthoryear{Ritter}{Ritter}{2012}]%
        {Ritter:2012aa}
\bibfield{author}{\bibinfo{person}{Dubravka Ritter}.}
  \bibinfo{year}{2012}\natexlab{}.
\newblock \bibinfo{title}{Do we still need the {E}qual {C}redit {O}pportunity
  {A}ct?}
\newblock
\newblock
\urldef\tempurl%
\url{https://ideas.repec.org/p/fip/fedpdp/12-03.html}
\showURL{%
\tempurl}


\bibitem[\protect\citeauthoryear{Ropiequet, Naveja, and Noonan}{Ropiequet
  et~al\mbox{.}}{2014}]%
        {Ropiequet:2014aa}
\bibfield{author}{\bibinfo{person}{John~L. Ropiequet},
  \bibinfo{person}{Christopher~S. Naveja}, {and} \bibinfo{person}{L.~Jean
  Noonan}.} \bibinfo{year}{2014}\natexlab{}.
\newblock \showarticletitle{Fair Lending Developments: Is Disparate Impact Here
  to Stay?}
\newblock \bibinfo{journal}{\emph{The Business Lawyer}} \bibinfo{volume}{69},
  \bibinfo{number}{2} (\bibinfo{year}{2014}), \bibinfo{pages}{609--621}.
\newblock
\urldef\tempurl%
\url{http://www.jstor.org/stable/43665748}
\showURL{%
\tempurl}


\bibitem[\protect\citeauthoryear{Rothstein}{Rothstein}{2018}]%
        {Rothstein:2018aa}
\bibfield{author}{\bibinfo{person}{Richard Rothstein}.}
  \bibinfo{year}{2018}\natexlab{}.
\newblock \bibinfo{booktitle}{\emph{The Color of Law: A Forgotten History of
  How Our Government Segregated America}}.
\newblock \bibinfo{publisher}{Liveright}.
\newblock


\bibitem[\protect\citeauthoryear{Rutherglen}{Rutherglen}{1987}]%
        {Rutherglen:1987aa}
\bibfield{author}{\bibinfo{person}{George Rutherglen}.}
  \bibinfo{year}{1987}\natexlab{}.
\newblock \showarticletitle{Disparate Impact under {Title VII}: An Objective
  Theory of Discrimination}.
\newblock \bibinfo{journal}{\emph{Virginia Law Review}} \bibinfo{volume}{73},
  \bibinfo{number}{7} (\bibinfo{year}{1987}), \bibinfo{pages}{1297--1345}.
\newblock
\urldef\tempurl%
\url{http://www.jstor.org/stable/1072940}
\showURL{%
\tempurl}


\bibitem[\protect\citeauthoryear{Steel and Angwin}{Steel and Angwin}{0 08}]%
        {Steel:2010aa}
\bibfield{author}{\bibinfo{person}{Emily Steel} {and} \bibinfo{person}{Julia
  Angwin}.} \bibinfo{year}{2010-08}\natexlab{}.
\newblock \showarticletitle{On the Web's Cutting Edge, Anonymity in Name Only}.
\newblock \bibinfo{journal}{\emph{Wall Street Journal}}
  (\bibinfo{year}{2010-08}).
\newblock
\urldef\tempurl%
\url{www.wsj.com/news/articles/SB10001424052748703294904575385532109190198}
\showURL{%
\tempurl}


\bibitem[\protect\citeauthoryear{Strader}{Strader}{5 06}]%
        {Strader:2015aa}
\bibfield{author}{\bibinfo{person}{Yolanda~P Strader}.}
  \bibinfo{year}{2015-06}\natexlab{}.
\newblock \bibinfo{title}{{CFPB} Continues Crackdown on Fair Lending: Marketing
  Materials Targeted}.
\newblock
\newblock
\urldef\tempurl%
\url{https://www.carltonfields.com/cfpb-continues-crackdown-on-fair-lending-marketing-materials-targeted/}
\showURL{%
\tempurl}


\bibitem[\protect\citeauthoryear{Turner}{Turner}{6 09}]%
        {Turner:2016aa}
\bibfield{author}{\bibinfo{person}{Ryan Turner}.}
  \bibinfo{year}{2016-09}\natexlab{}.
\newblock \showarticletitle{A model explanation system}. In
  \bibinfo{booktitle}{\emph{2016 IEEE 26th International Workshop on Machine
  Learning for Signal Processing (MLSP)}}. \bibinfo{pages}{1--6}.
\newblock
\urldef\tempurl%
\url{https://doi.org/10.1109/MLSP.2016.7738872}
\showDOI{\tempurl}


\bibitem[\protect\citeauthoryear{{US Congress}}{{US Congress}}{1968a}]%
        {fha}
\bibfield{author}{\bibinfo{person}{{US Congress}}.}
  \bibinfo{year}{1968}\natexlab{a}.
\newblock \bibinfo{title}{{42 U.S.C. §3601 ff.: Fair Housing Act}}.
\newblock
\newblock
\urldef\tempurl%
\url{https://www.justice.gov/crt/fair-housing-act-2}
\showURL{%
\tempurl}


\bibitem[\protect\citeauthoryear{{US Congress}}{{US Congress}}{1968b}]%
        {ccpa}
\bibfield{author}{\bibinfo{person}{{US Congress}}.}
  \bibinfo{year}{1968}\natexlab{b}.
\newblock \bibinfo{title}{{Consumer Credit Protection Act}}.
\newblock
\newblock
\urldef\tempurl%
\url{http://uscode.house.gov/view.xhtml?path=/prelim@title15/chapter41&edition=prelim}
\showURL{%
\tempurl}


\bibitem[\protect\citeauthoryear{{US Congress}}{{US Congress}}{0 10}]%
        {fcra}
\bibfield{author}{\bibinfo{person}{{US Congress}}.}
  \bibinfo{year}{1970-10}\natexlab{}.
\newblock \bibinfo{title}{{15 U.S.C. §1681 ff.: Fair Credit Reporting Act}}.
\newblock
\newblock
\urldef\tempurl%
\url{https://www.gpo.gov/fdsys/pkg/STATUTE-84/pdf/STATUTE-84-Pg1114-2.pdf}
\showURL{%
\tempurl}


\bibitem[\protect\citeauthoryear{{US Congress}}{{US Congress}}{4 10}]%
        {ecoa}
\bibfield{author}{\bibinfo{person}{{US Congress}}.}
  \bibinfo{year}{1974-10}\natexlab{}.
\newblock \bibinfo{title}{{15 U.S.C. §1691 ff.: Equal Credit Opportunity
  Act}}.
\newblock
\newblock
\urldef\tempurl%
\url{https://www.ecfr.gov/cgi-bin/text-idx?tpl=/ecfrbrowse/Title12/12cfr202_main_02.tpl}
\showURL{%
\tempurl}


\bibitem[\protect\citeauthoryear{{US Congress}}{{US Congress}}{3 12}]%
        {facta}
\bibfield{author}{\bibinfo{person}{{US Congress}}.}
  \bibinfo{year}{2003-12}\natexlab{}.
\newblock \bibinfo{title}{{P. L. 108-159: Fair and Accurate Credit Transactions
  Act of 2003}}.
\newblock
\newblock
\urldef\tempurl%
\url{https://www.gpo.gov/fdsys/pkg/PLAW-108publ159/pdf/PLAW-108publ159.pdf}
\showURL{%
\tempurl}


\bibitem[\protect\citeauthoryear{{US Congress}}{{US Congress}}{7 03}]%
        {feha}
\bibfield{author}{\bibinfo{person}{{US Congress}}.}
  \bibinfo{year}{2017-03}\natexlab{}.
\newblock \bibinfo{title}{Fair and Equal Housing Act of 2017}.
\newblock
\newblock
\urldef\tempurl%
\url{https://www.congress.gov/bill/115th-congress/house-bill/1447}
\showURL{%
\tempurl}


\bibitem[\protect\citeauthoryear{{US Court of Appeals for the Eighth
  Circuit}}{{US Court of Appeals for the Eighth Circuit}}{3 03}]%
        {Taylor:2013}
\bibfield{author}{\bibinfo{person}{{US Court of Appeals for the Eighth
  Circuit}}.} \bibinfo{year}{2013-03}\natexlab{}.
\newblock \bibinfo{title}{No. 11-3466: {C}atherine {L}. {T}aylor v. {T}enant
  {T}racker, {I}nc.}
\newblock
\newblock
\urldef\tempurl%
\url{https://www.gpo.gov/fdsys/pkg/USCOURTS-ca8-11-03648/pdf/USCOURTS-ca8-11-03648-0.pdf}
\showURL{%
\tempurl}


\bibitem[\protect\citeauthoryear{Vermorel and Mohri}{Vermorel and
  Mohri}{2005}]%
        {Vermorel:2005aa}
\bibfield{author}{\bibinfo{person}{Joann\`es Vermorel} {and}
  \bibinfo{person}{Mehryar Mohri}.} \bibinfo{year}{2005}\natexlab{}.
\newblock \showarticletitle{Multi-armed Bandit Algorithms and Empirical
  Evaluation}. In \bibinfo{booktitle}{\emph{Machine Learning: ECML 2005}},
  \bibfield{editor}{\bibinfo{person}{Jo\~ao Gama}, \bibinfo{person}{Rui
  Camacho}, \bibinfo{person}{Pavel~B. Brazdil},
  \bibinfo{person}{Al\'ipio~M\'ario Jorge}, {and} \bibinfo{person}{Lu\'is
  Torgo}} (Eds.). \bibinfo{publisher}{Springer Berlin Heidelberg},
  \bibinfo{pages}{437--448}.
\newblock


\bibitem[\protect\citeauthoryear{Wei, Yildirim, den Bulte, and Dellarocas}{Wei
  et~al\mbox{.}}{2016}]%
        {Wei:2016aa}
\bibfield{author}{\bibinfo{person}{Yanhao Wei}, \bibinfo{person}{Pinar
  Yildirim}, \bibinfo{person}{Christophe~Van den Bulte}, {and}
  \bibinfo{person}{Chrysanthos Dellarocas}.} \bibinfo{year}{2016}\natexlab{}.
\newblock \showarticletitle{Credit Scoring with Social Network Data}.
\newblock \bibinfo{journal}{\emph{Marketing Science}} \bibinfo{volume}{35},
  \bibinfo{number}{2} (\bibinfo{year}{2016}), \bibinfo{pages}{234--258}.
\newblock
\urldef\tempurl%
\url{https://doi.org/10.1287/mksc.2015.0949}
\showDOI{\tempurl}


\bibitem[\protect\citeauthoryear{Zhang}{Zhang}{6 01}]%
        {Zhang:2016aa}
\bibfield{author}{\bibinfo{person}{Yan Zhang}.}
  \bibinfo{year}{2016-01}\natexlab{}.
\newblock \showarticletitle{Assessing Fair Lending Risks Using Race/Ethnicity
  Proxies}.
\newblock \bibinfo{journal}{\emph{Management Science}} \bibinfo{volume}{64},
  \bibinfo{number}{1} (\bibinfo{year}{2016-01}), \bibinfo{pages}{178--197}.
\newblock
\urldef\tempurl%
\url{https://doi.org/10.1287/mnsc.2016.2579}
\showDOI{\tempurl}


\bibitem[\protect\citeauthoryear{Zimmer}{Zimmer}{1996}]%
        {Zimmer:1996aa}
\bibfield{author}{\bibinfo{person}{Michael~J Zimmer}.}
  \bibinfo{year}{1996}\natexlab{}.
\newblock \showarticletitle{The Emerging Uniform Structure of Disparate
  Treatment Discrimination Litigation}.
\newblock \bibinfo{journal}{\emph{Georgia Law Review}}  \bibinfo{volume}{30}
  (\bibinfo{year}{1996}), \bibinfo{pages}{563--626}.
\newblock
\urldef\tempurl%
\url{https://ssrn.com/abstract=1354323}
\showURL{%
\tempurl}


\end{thebibliography}

\end{document}